\documentclass[letterpaper, 10 pt, conference]{ieeeconf}  
\newcommand{\ElsIf}[1]{\textbf{else if}\ (#1)}

\IEEEoverridecommandlockouts                              

\usepackage[nospace]{cite}
\usepackage{amsmath,amssymb,amsfonts}
\usepackage{graphicx}
\usepackage{textcomp}
\usepackage[dvipsnames]{xcolor}
\usepackage{multirow}
\usepackage{makecell}
\usepackage{tikz}
\usepackage{color, colortbl}
\usepackage{xcolor}
\usepackage{url}
\definecolor{LightBlue}{RGB}{212, 250, 252} 
\usepackage[flushleft]{threeparttable}
\usepackage{booktabs}
\usetikzlibrary{arrows.meta}
\let\labelindent\relax
\usepackage{enumitem}
\definecolor{LightBlue}{RGB}{212, 250, 252}
\usepackage{caption}
\usepackage{subcaption}
\usepackage{tabulary}
\usepackage{algpseudocode}
\usepackage{amsmath}
\usepackage[utf8]{inputenc}
\usepackage[T1]{fontenc}
\def\mycircle[#1]{\tikz\draw[#1,fill=#1] (0,0) circle (0.125cm);}

\def\arrowright[#1]{\begin{tikzpicture}
  \draw[#1, -{Triangle[width = 5pt, length = 2pt]}, line width = 2pt] (0.0, 0.0) -- (0.25, 0.0);\path (current bounding box.south west) +(0,-0.06);
\end{tikzpicture}}

\def\arrowleft[#1]{\begin{tikzpicture}
  \draw[#1, -{Triangle[width = 5pt, length = 2pt]}, line width = 2pt] (0.25, 0.0) -- (0.0, 0.0);\path (current bounding box.south west) +(0,-0.06);

\end{tikzpicture}}

\def\arrowup[#1]{\begin{tikzpicture}
  \draw[#1, -{Triangle[width = 5pt, length = 2pt]}, line width = 2pt, rotate=270] (0.25, 0.0) -- (0.0, 0.0);
\end{tikzpicture}}

\def\arrowstraightleft[#1]{\begin{tikzpicture}
\draw[#1, -{Triangle[width = 5pt, length = 2pt]}, line width = 2pt, rotate=270] (0.25, 0.0) -- (0.0, 0.0);
  \draw[#1, -{Triangle[width = 5pt, length = 2pt]}, line width = 2pt] (0.0, -0.17) -- (-0.15, -0.17);
\end{tikzpicture}}

\def\arrowstraightright[#1]{\begin{tikzpicture}
    \draw[#1, -{Triangle[width = 5pt, length = 2pt]}, line width = 2pt, rotate=270] (0.25, 0.0) -- (0.0, 0.0);

  \draw[#1, -{Triangle[width = 5pt, length = 2pt]}, line width = 2pt] (0.0, -0.17) -- (0.15, -0.17);

\end{tikzpicture}}

\definecolor{Pistachio}{RGB}{140, 212, 126} 
\definecolor{CrayolaYellow}{RGB}{248, 214, 109}
\definecolor{PastelRed}{RGB}{255, 105, 97}
\definecolor{PastelOrange}{RGB}{255, 181, 76}
\definecolor{SoftCharcoal}{RGB}{66,66, 66}
\title{\LARGE \bf Centralized Decision-Making for Platooning By Using SPaT-Driven Reference Speeds}

\author{%
  Melih Yazgan\textsuperscript{1*}, 
  S\"uleyman Tatar\textsuperscript{3*} and 
  J. Marius Z\"ollner\textsuperscript{1,2}%
  \thanks{\textsuperscript{1}FZI Research Center for Information Technology, Dept.\ of Technical Cognitive Systems, Haid‑und‑Neu Str.~10–14, Karlsruhe, Germany. \texttt{\{surname\}@fzi.de}}%
  \thanks{\textsuperscript{2}Karlsruhe Institute of Technology (KIT), Institute of Applied Technical Cognitive Systems, Kaiserstr.~12, Karlsruhe, Germany. \texttt{\{prename.surname\}@kit.edu}}%
  \thanks{\textsuperscript{3}Karlsruhe Institute of Technology (KIT), Institute for Information Processing Technology, Engesserstr.~5, Karlsruhe, Germany. \texttt{\{prename.surname\}@student.kit.edu}}%
  \thanks{\textsuperscript{*}Equal contribution}%
}

\begin{document}

\maketitle
\thispagestyle{empty}
\pagestyle{empty}

\begin{abstract} 
This paper introduces a centralized approach for fuel-efficient urban platooning by leveraging real-time Vehicle-to-Everything (V2X) communication and Signal Phase and Timing (SPaT) data. A nonlinear Model Predictive Control (MPC) algorithm optimizes the trajectories of platoon leader vehicles, employing an asymmetric cost function to minimize fuel-intensive acceleration. Following vehicles utilize a gap- and velocity-based control strategy, complemented by dynamic platoon splitting logic communicated through Platoon Control Messages (PCM) and Platoon Awareness Messages (PAM). Simulation results obtained from the CARLA environment demonstrate substantial fuel savings of up to 41.2\%, along with smoother traffic flows, fewer vehicle stops, and improved intersection throughput.
\end{abstract}

\section{Introduction}
\label{sec:introduction}
Transportation emissions significantly contribute to environmental challenges, necessitating innovative solutions for eco-friendly urban mobility \cite{EuropeanCommission2016}. Urban areas, characterized by frequent stops, idling, and congestion, are major contributors to energy inefficiency, making sustainable driving strategies increasingly essential. Eco-driving practices, such as maintaining steady speeds, anticipating traffic signal changes, and minimizing abrupt acceleration or deceleration, have emerged as key approaches to address these challenges.

Platooning, coordinating a group of vehicles to drive closely together, represents a particularly promising solution for eco-driving in urban scenarios due to its potential to reduce aerodynamic drag, decrease vehicle-to-vehicle spacing, and harmonize acceleration and braking patterns across multiple vehicles. These coordinated behaviors enable smoother traffic flows and reduced stop-and-go events, which directly translate into substantial fuel savings and emissions reductions. Furthermore, platooning can leverage real-time traffic data, such as SPaT information, more effectively by synchronizing multiple vehicles' speeds and trajectories, enhancing the overall efficiency of the urban transportation system.

However, the recent advent of connected vehicle technologies and advanced predictive control mechanisms has opened new opportunities for implementing eco-driving principles in urban settings. In particular, V2X communication has been instrumental in advancing research aimed at optimizing urban traffic systems. A notable development within this domain is the use of SPaT data, which provides real-time information about the current state (green, amber, red) and precise timing (remaining durations) of traffic signals. By accessing SPaT data, vehicles can anticipate future signal states and optimize their speed profiles accordingly, thus significantly reducing unnecessary acceleration, deceleration, and idle times typically experienced at urban intersections. This predictive approach directly translates into notable reductions in fuel consumption and emissions.

\section{Related Work}
Urban platooning and eco-driving are widely studied as strategies to reduce fuel consumption and improve traffic flow. Prior research generally falls into decentralized and centralized control paradigms. Decentralized methods empower vehicles with local decision-making capabilities, often using Vehicle-to-Vehicle (V2V) communication, but can face challenges in achieving optimal coordination, particularly at intersections. Centralized strategies, conversely, utilize a leader or central controller to manage the platoon, potentially offering better overall coordination. This section reviews key works within both approaches, highlighting their respective strengths and limitations concerning fuel-efficient urban platooning.

\subsection{Decentralized Platoon Management Approaches}
Decentralized control allows individual vehicles to make local decisions while communicating to enhance efficiency and safety. For instance, Hu et al.~\cite{Hu2022} investigated a switching-based MPC scheme where each vehicle independently optimizes its fuel consumption via V2V, achieving approximately 6.84\% fuel savings over conventional tracking methods in simulations. Extending beyond basic peer-to-peer setups, Schmied et al.~\cite{schmied2015nonlinear} developed a nonlinear MPC-based Cooperative Adaptive Cruise Control (CACC) system integrating infrastructure-to-vehicle data. By fusing radar signals with SPaT information, their approach allows vehicles to anticipate lead vehicle behavior and adapt spacing, penalizing constraint violations to maintain safety and efficiency.

Other research has focused specifically on urban fuel optimization. HomChaudhuri et al.~\cite{homchaudhuri_fast_2017} applied a decentralized, fast MPC formulation using V2V and SPaT data to adjust speed profiles and minimize idling time at red lights. Their use of a linearized MPC model supports computationally efficient real-time implementation. Similarly, Zhao et al.~\cite{zhao2018CoDrive} introduced CoDrive, a cooperative decentralized system optimizing speeds near urban intersections. CoDrive uses V2V communication to negotiate a shared speed among vehicles potentially diverging later, aiming to reduce overall fuel consumption and idling by incorporating SPaT data into an MPC routine. Simulation results in SUMO~\cite{SUMO2018} indicated fuel savings up to 38.2\% compared to baseline driving and ~7.9\% over non-cooperative advisories. Complementing these efforts, Chada et al.~\cite{chadaEcologicalAdaptiveCruise2020} propose a system for urban driving that employs two linear MPC controllers, one to optimize speed for green‑wave coordination when the road is clear, and the other to maintain energy‑efficient, safe car‑following. The system switches between these modes using real‑time sensor data and intersection proximity, achieving up to 26\% energy savings in green‑wave tracking and 13\% during car‑following.

Despite these advances, decentralized methods face limitations. A primary challenge is the lack of inherent centralized coordination needed to fully exploit real-time SPaT data for platoon-wide optimization at intersections, evident in approaches like~\cite{Hu2022, schmied2015nonlinear}. Furthermore, strategies relying purely on local decision-making~\cite{homchaudhuri_fast_2017, zhao2018CoDrive} can sometimes lead to abrupt maneuvers and typically lack mechanisms for dynamic platoon splitting when only a subset of vehicles can pass an intersection, potentially reducing overall throughput and efficiency.

\subsection{Centralized Platoon Management Approaches}
In contrast, centralized platoon management employs a central controller or a designated leader to manage the entire group's motion. Asadi and Vahidi~\cite{asadiPredictiveCruiseControl2011}, for example, designed a predictive cruise control framework integrating traffic signal data into adaptive cruise control. Their MPC-based optimization calculates velocity profiles synchronized with green lights to minimize stops, significantly reducing fuel consumption and brake wear. Kennedy et al.~\cite{kennedy2023centralized} explored a centralized MPC design that maintains string stability and safety even during brief periods of human driver intervention. By incorporating driver preferences and predicted trajectories, the system coordinates the platoon despite temporary manual control inputs.

Focusing on signalized intersections, Ma et al.~\cite{Ma2021EcoCACC} put forward a centralized Eco-CACC system merging eco-driving principles with CACC. A leader determines an optimal velocity trajectory using dynamic programming, while followers use PID-based control to maintain headway. The system adapts its objective based on signal phases to facilitate green light passage, demonstrating 8.02\% energy savings over manual driving in experiments. Lin et al.~\cite{ecoDrivingMultipleSignalizedIntersections} offered another centralized eco-driving solution for traversing multiple signalized intersections, formulating an open-loop optimal control problem using detailed powertrain models. Their multi-stage driving rule, guided by Dijkstra's algorithm, reduces computational complexity while seeking energy-efficient trajectories.
Han et al.~\cite{hanLeveragingMultipleConnected2021} propose a centralized, hierarchical V2I-based speed planner with two key modules: a green window selector that uses Dijkstra’s algorithm to pick optimal green‑light sequences, and a reference trajectory generator that calculates energy‑minimizing speed profiles across intersections. Simulations showed substantial energy savings and smoother speed profiles versus decentralized methods.

However, centralized strategies also exhibit drawbacks. Some frameworks rely heavily on predictive or precomputed signal data, potentially limiting real-time adaptability~\cite{asadiPredictiveCruiseControl2011}. Others may maintain coordination during interventions but lack explicit mechanisms for dynamic platoon splitting when needed~\cite{kennedy2023centralized}. Additionally, simpler follower control models like PID might not fully capture complex dynamics~\cite{Ma2021EcoCACC}, while computationally efficient open-loop approaches can struggle with unexpected signal timing fluctuations~\cite{ecoDrivingMultipleSignalizedIntersections}.

Addressing the identified gaps in both decentralized approaches (e.g., intersection coordination, dynamic splitting) and centralized strategies (e.g., real-time adaptability, splitting capabilities, follower dynamics), this research introduces a centralized platoon management framework implemented within the open-source CARLA simulation environment. The frameworSk features a nonlinear MPC controller for the platoon leader to optimize maneuvers at intersections, coupled with a gap-and-velocity-based control strategy for following vehicles that utilizes real-time V2V and SPaT data. Crucially, dynamic platoon splitting logic is integrated to enable subgroups to proceed independently when full group passage is infeasible, thereby enhancing intersection throughput and reducing idle times.

\section{Methodology}
\label{sec:methodology}

This section outlines the proposed methodology for optimizing fuel consumption and travel efficiency in connected vehicle platoons using a centralized \textit{pack leader scenario}. The pack leader dynamically adjusts its velocity based on SPaT data guiding the following vehicles to maintain cohesive and energy-efficient platoon behavior. To achieve this, the pack leader employs an MPC, while the following vehicles utilize PID controllers for longitudinal and lateral control. The following subsections detail the design and implementation of these control strategies and scenarios

\subsection{Simulation Framework in CARLA}
\begin{figure*}[h]
    \centering
    \includegraphics[width=12cm, height=6cm]{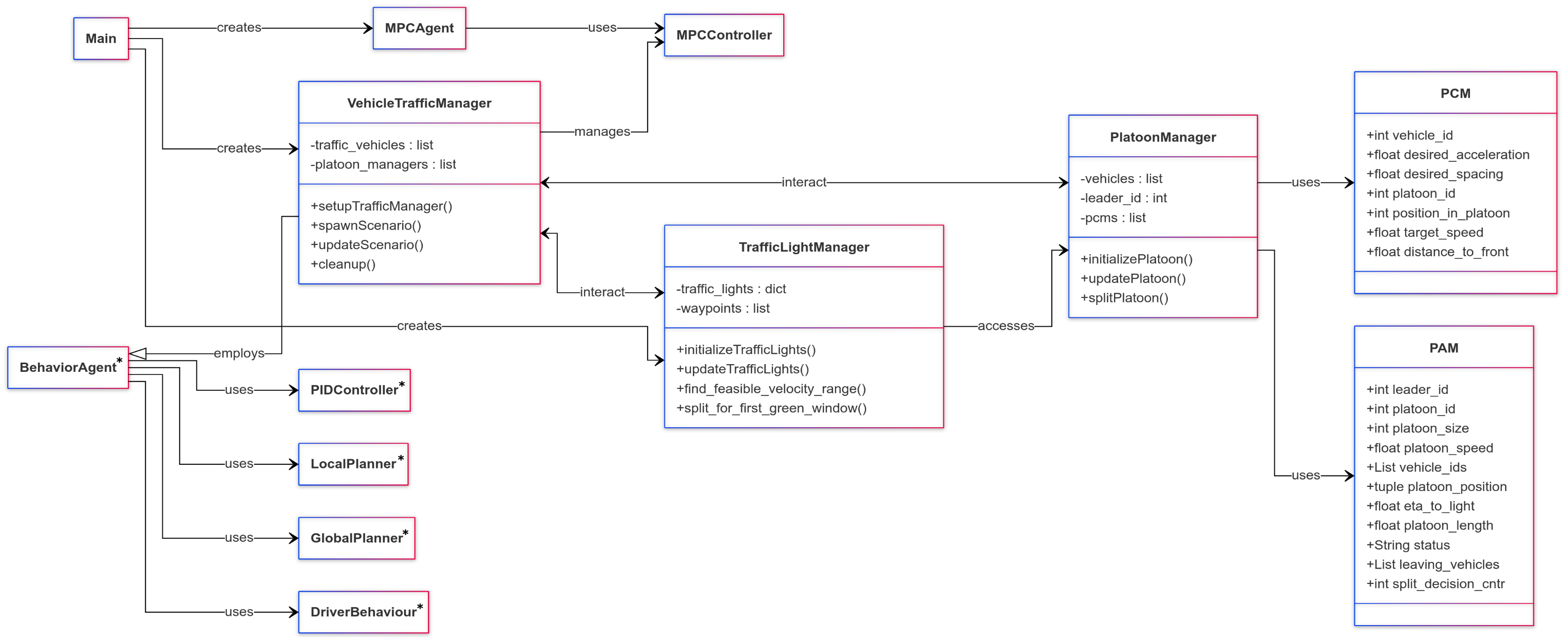}
    \caption{System architecture of the CARLA platooning framework, showing vehicle control, traffic signal integration, and V2X messaging for coordinated platoon management.}
    \label{fig:carla_sim}
    {\footnotesize The packages marked with an asterisk (*) are taken from CARLA API examples and have been improved based on specific requirements.}
\end{figure*}

\begin{figure}[h]
    \centering
    \includegraphics[width=6cm, height=4cm]{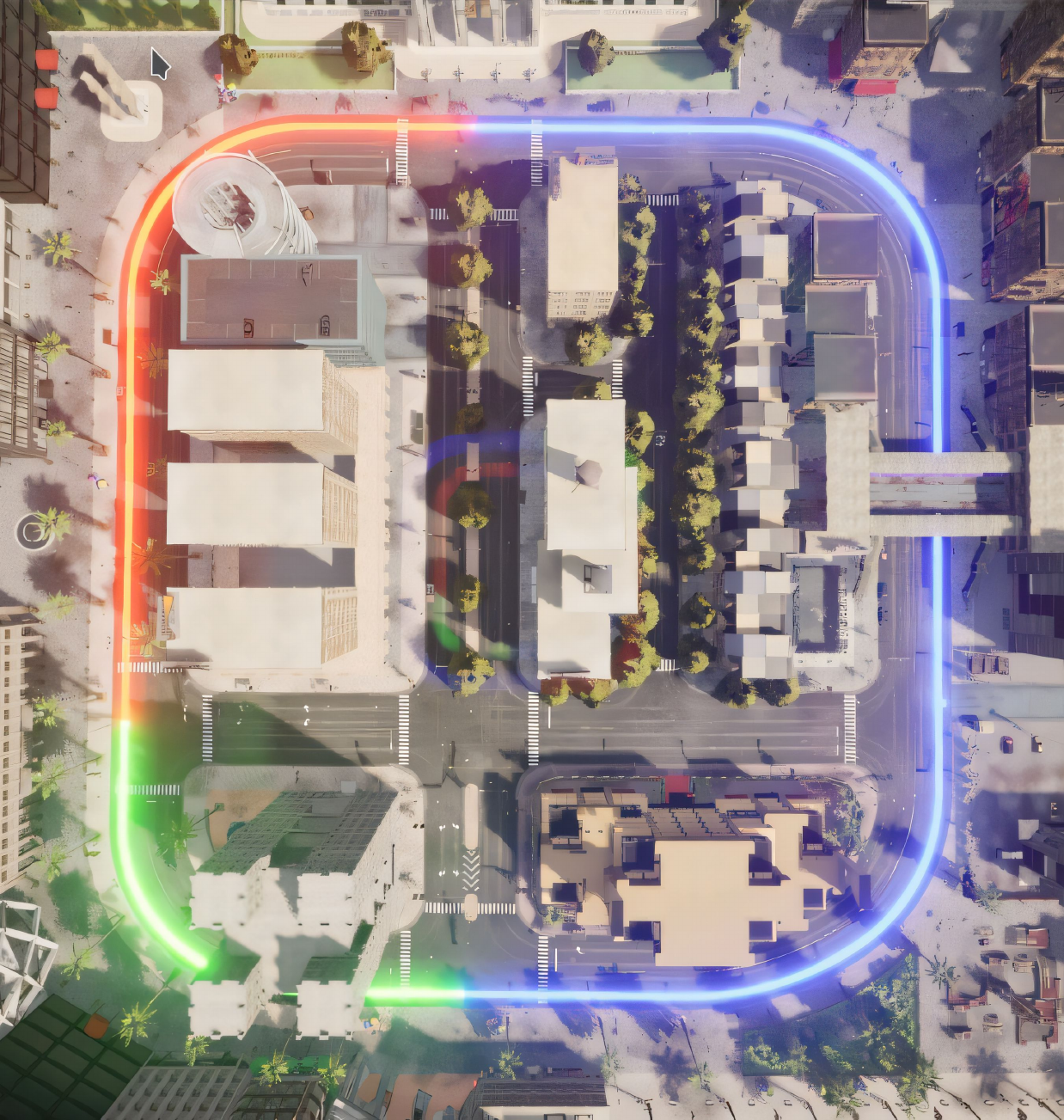}
    \caption{Simulation environment in CARLA showing the test route with three signalized intersections.The colored paths represent different traffic corridors: Corridor 1 (red), Corridor 2 (green), and Corridor 3 (blue).}
    \label{fig:carla_route}
\end{figure}

A custom simulation framework in CARLA~\cite{dosovitskiy2017carla} has been developed to evaluate the pack leader scenario and green window advisory strategy with platoon split decision logic. It integrates a nonlinear MPC for the pack leader and PID controllers for follower vehicles, incorporating a green window advisory algorithm that utilizes SPaT data and supports V2X messaging (Platoon PAM and PCM). Operating at 10 Hz, it ensures precise trajectory planning and smooth intersection transitions. The 800m circular route features three traffic lights, with segments defined as Corridor 1 (red), Corridor 2 (green), and Corridor 3 (blue), as illustrated in Figure~\ref {fig:carla_route}. The framework includes modules for vehicle control, platoon management, and traffic light interaction. The VehicleTrafficManager handles traffic and platoon operations, while the TrafficLightManager updates signals based on SPaT messages. The MPCAgent controls the pack leader using the MPCController, while follower vehicles use the BehaviorAgent with a PIDController. Path planning is managed by LocalPlanner and GlobalPlanner, and DriverBehaviour handles vehicle reactions. The PlatoonManager tracks the platoon, managing PCMs and PAMs to synchronize decision-making. PAM messages, based on ETSI TR 103 429 V2.1.1 and the ENSEMBLE project~\cite{ENSEMBLE2022}, convey leader ID, platoon size, arrival time, and split decisions. PlatoonManager determines split decisions for optimized green window usage, balancing intersection efficiency and platoon coherence. The nonlinear MPC, implemented in CasADi~\cite{Andersson2019}, operates on a 5-second horizon, synchronized with the CARLA environment. Code and supplementary materials will be available at \underline{https://url.fzi.de/EcoLead}.

\subsection{MPC Design}
\label{sec:mpc}
The MPC predicts the future state of the vehicle over a finite time horizon ($N$) and computes optimal control inputs to minimize a carefully designed cost function. The vehicle's state vector is defined as:
\begin{equation}
\mathbf{s} =
\begin{bmatrix}
x & y & \psi & v & \text{CTE} & \epsilon_\psi
\end{bmatrix}
\end{equation}

Where:
\begin{itemize}
    \item $x, y$: Position coordinates.
    \item $\psi$: Orientation.
    \item $v$: Longitudinal velocity.
    \item $\text{CTE}$: Cross-track error, indicating the lateral deviation from the desired trajectory.
    \item $\epsilon_\psi$: Orientation error, representing the angular misalignment.
\end{itemize}The state variables include the vehicle's position $(x, y)$, heading angle $(\psi)$, longitudinal velocity $(v)$, and control inputs $(u, \delta)$. The system dynamics are governed by:
\begin{align}
x_{t+1} &= x_t + v_t \cos(\psi_t) \Delta t, \\
y_{t+1} &= y_t + v_t \sin(\psi_t) \Delta t, \\
\psi_{t+1} &= \psi_t + \frac{v_t \delta_t}{L_f} \Delta t, \\
v_{t+1} &= v_t + a_{net} \Delta t, \\
\text{CTE}_{t+1} &= \text{CTE}_t + v_t \sin(\epsilon_{\psi,t}) \Delta t, \\
\epsilon_{\psi, t+1} &= \epsilon_{\psi,t} + \frac{v_t \delta_t}{L_f} \Delta t
\end{align}
Where $L_f$ represents the distance from the front axle to the vehicle's center of gravity and $a_{net}$ is the acceleration in which we take the rolling and air resistance drag into the state equations. The parameters are taken from the work of Chen et al.~\cite{chen2021learning}, in which the bicycle model is tuned for CARLA.
\begin{equation}
    a_{\text{net}} = u - \frac{0.5 C_D \rho_a A_v v^2}{M_h} - \mu g
    \end{equation}
where:
\begin{itemize}
    \item $u:$ is the traction or braking force per unit mass in m/s²,
    \item $\delta:$ is the steering control input,
    \item $C_D:$ is the drag coefficient,
    \item $\rho_a:$ is the air density in kg/m³,
    \item $A_v:$ is the frontal area of the vehicle in m²,
    \item $M_h:$ is the vehicle's mass in kg,
    \item $\mu:$ is the rolling resistance coefficient,
    \item $g:$ is the gravitational acceleration m/s².
\end{itemize}

\paragraph{Cost Function}
The cost function reflects the fuel efficiency objectives utilizing the reference velocity, $v_{\text{ref}}$, from SPaT-based speed advisory and tracking. Additionally, it penalizes deviations from the
reference trajectory, actuator usage, and abrupt changes in
control inputs. The cost function is designed as:
\begin{itemize}
    \item \textbf{Deviation from the reference trajectory}:
    \begin{equation}
    \text{Cost}_{\text{tracking}} = \sum_{t=0}^{N} \left( w_{\text{cte}} \cdot \text{CTE}_t^2 + w_{\epsilon} \cdot \epsilon_{\psi,t}^2 \right)
\end{equation}
    \item \textbf{Deviation from the reference velocity}:
    \begin{equation}
    \text{Cost}_{\text{velocity}} = \sum_{t=0}^{N} w_{v} (v_{\text{ref}} - v_t)^2
    \end{equation}

    \item \textbf{Actuator usage}:
    \begin{align}
    \text{Cost}_{\text{actuator}} &= \sum_{t=0}^{N-2} \big( w_{\Delta\delta} (\delta_{t+1} - \delta_t)^2 \nonumber \\
    &\quad + w_{\Delta a} (a_{t+1} - a_t)^2 \big)
    \end{align}
\end{itemize}

\paragraph{Constraints}
The optimization problem is subject to the following constraints:
\begin{align}
    v_{\text{min}} &\leq v_t \leq v_{\text{max}}, \quad \forall t, \\
    a_{\text{min}} &\leq a_t \leq a_{\text{max}}, \quad \forall t, \\
    \delta_{\text{min}} &\leq \delta_t \leq \delta_{\text{max}}, \quad \forall t, \\
    \text{CTE}_{t+1} &= f(x_{t+1}) - y_{t+1}, \quad \forall t, \\
    \epsilon_{\psi t+1} &= \psi_{t+1} - \psi_{\text{des}}, \quad \forall t.
\end{align}

\subsection{Control for Vehicles in Platoon}

\label{sec:car_following}
Cooperative Adaptive Cruise Control (CACC) enhances vehicle-following capabilities by incorporating V2V communication and gap-based velocity control. Unlike traditional cruise control approaches, CACC dynamically adjusts velocity using an adaptive PID-based strategy. This ensures smooth platoon behavior, minimizes velocity oscillations, and maintains stable inter-vehicle gaps.
The CACC module consists of two primary components:

\begin{enumerate}
    \item \textbf{Gap-Based Velocity Controller (CACC PID)}: Computes a target velocity based on the inter-vehicle gap and preceding vehicle and platoon leader velocity. Adaptive gain scheduling ensures robust control across varying speed conditions.
    \item \textbf{Throttle/Brake Execution}: The velocity output from CACC is passed to CARLA's internal PID controller, translating it into appropriate throttle and brake commands.
\end{enumerate}

The CACC module computes a target velocity \( v_{\text{target}} \) based on the preceding vehicle's speed and the inter-vehicle gap error:
\begin{equation}
    v_{\text{target}} = v_{\text{preceding}} + K_1(s_{\text{current}} - s_{\text{desired}}) + K_2 (v_{\text{leader}} - v_{\text{preceding}})
\end{equation}

\vspace{3cm}
where:
\begin{itemize}
    \item \( v_{\text{preceding}} \) is the velocity of the preceding vehicle,
    \item \( v_{\text{leader}} \) is the velocity of the MPC-controlled leader vehicle,
    \item \( s_{\text{desired}} \) is the dynamically computed desired inter-vehicle gap:
    The desired gap $s_{\text{desired}}$ is dynamically computed as:
    \begin{equation}
        s_{\text{desired}} = \max(s_{\min},\, \min(s_0 + v_{\text{ego}} T_{\text{gap}},\, s_{\max}))
    \end{equation}
where $s_{\min} = 10$ m, $s_{\max} = 15$ m, $s_0 = 1$ m (standstill gap), and $T_{\text{gap}} = 1$ s.
    \item \( s_{\text{current}} \) is the actual measured gap based on V2V,
    \item $K_1$ and $K_2$ are adaptive gains computed as shown below. The maximum and minimum values are manually fine-tuned:  
\[
K_1 \in \{0.2, 0.6\}, \quad K_2 \in \{0.3, 0.7\}.
\]
    \footnotesize
    \begin{equation}
        K_1 = K_{1_{\min}} + (K_{1_{\max}} - K_{1_{\min}}) \times (1 - e^{-\frac{\text{gap error}}{s_{\max}}})
    \end{equation}
    \begin{equation}
        K_2 = K_{2_{\min}} + (K_{2_{\max}} - K_{2_{\min}}) \times (1 - e^{-\frac{\text{gap error}}{s_{\max}}})
    \end{equation}
\end{itemize}
This formulation ensures smooth adjustments to \( v_{\text{target}} \) without excessive fluctuations.
A PID controller regulates the ego vehicle's velocity using:

\begin{equation}
    \text{speed error} = v_{\text{target}} - v_{\text{ego}}
\end{equation}

The PID gains are dynamically adjusted for different speed ranges and manually finetuned.

\subsection{Fuel Consumption Model}
\label{sec:fuel_consump}
The proposed fuel consumption model is a regression-based approach inspired by the work of Kamal et al. \cite{kamal2011ecological}. It computes the instantaneous fuel consumption of a vehicle as a function of its speed $v$ in m/s and acceleration $a$ in m/s², producing an output in Liters per second (L/s). This model simplifies the inherently complex process of fuel consumption, which depends on numerous factors such as engine speed, torque, temperature, weather and road conditions, velocity, acceleration, and others\cite{homchaudhuriFuelEconomicModel2015}.
\begin{equation}
    f_{\text{cruise}} = b_0 + b_1 v + b_2 v^2 + b_3 v^3,
\end{equation}
\begin{equation}
    f_{\text{accel}} = a_{\text{net}}(c_0 + c_1 v + c_2 v^2),
\end{equation}
\begin{equation}
    \zeta = 
    \begin{cases} 
    1 & \text{if } v \approx 0 \text{ or } u < 0, \\
    0 & \text{otherwise.}
    \end{cases}
\end{equation}
\begin{equation}
    \text{Fuel Consumption} = (1 - \zeta)(f_{\text{cruise}} + f_{\text{accel}}) + \zeta  f_{\text{id}}
\end{equation}
where the coefficients \(b_0\), \(b_1\), \(b_2\), \(b_3\), \(c_0\), \(c_1\), and \(c_2\) are polynomial coefficients specific to the vehicle and taken from \cite{kamal2011ecological}. Here, \( f_{\text{id}} \) represents the fuel consumption rate when the vehicle is stationary or braking.

\subsection{Speed Advisory Algorithm for Platoon with Split Decision}
\label{sec:algo}
The algorithm optimizes the reference velocity ($v_{\text{ref}}$) of a vehicle platoon approaching a traffic signal to ensure efficient and safe passage within the available green window ($[t_g^{\text{start}}, t_g^{\text{end}}]$). The algorithm operates through five key phases, each corresponding to specific lines in the pseudocode, as demonstrated in Algorithm 1.

\begin{itemize}
\item \textbf{Feasibility Assessment of Velocities (Line 14):}\\  
Once the initial delay has passed, the algorithm determines the feasible velocity bounds by calling \texttt{GetFeasibleVelocities} (Line 14). This function computes the lower (\(v_{\min}\)) and upper (\(v_{\max}\)) velocity limits based on the distance to the traffic light (\(d_{\text{tl}}\)) and the green phase window (\(t_g^{\text{start}}\) to \(t_g^{\text{end}}\)) received from SPaT data. (If SPaT is not yet available, typically until the leader is within 300\,m due to Dedicated Short Range Communications (DSRC) limitations\cite{Garcia2021Tutorial}, the platoon continues with its previous velocity.)

\item \textbf{Saturation Flow Verification (Lines 15-18):}\\  
Next, the algorithm checks if the platoon formation itself can clear the intersection distance within the current green window without requiring excessive speed. This is done via the \texttt{SaturationCheck} function (Line 16). This check ensures that the speed required for the platoon's length to pass through the intersection within the available green time if the leader doesn't pass the traffic light already, calculated as
\[
 v_{\text{required}} = \frac{d_\text{platoon}}{t_{\text{green\_window}}},
\]
does not exceed the maximum feasible velocity \(v_{\max}\) determined earlier. PAM Message has already \(d_\text{platoon}\), calculated from each trailing vehicles PCM. If \(v_{\text{required}}\) is too high (i.e., it exceeds \(v_{\max}\)), saturation is considered unachievable for the entire group (\texttt{False}). In that case, the algorithm mandates a platoon split by calling \texttt{SplitPlatoon} (Line 17) and then adjusts the front subgroup's velocity with \texttt{AdjustForSplit} (Line 18) so that the vehicles in the front group can pass within the green window.
\item \textbf{Optimization and Arrival Time Checks (Lines 19-28):}\\  
If the saturation check passes, the algorithm proceeds to optimize the platoon's velocity within the feasible range using \texttt{OptimizeVelocity} (Line 20), which yields a candidate velocity \(v_{\text{cand}}\).  
If both \(t_{\text{lead}}\) and \(t_{\text{last}}\) are within the green window (i.e., \(t_g^{\text{start}} \leq t_{\text{lead}} \leq t_g^{\text{end}}\) and \(t_{\text{last}} \leq t_g^{\text{end}}\)) (Line 21), the algorithm returns \(v_{\text{cand}}\) along with the entire platoon (Line 23). Otherwise, if the candidate velocity does not allow the whole platoon to pass in time, the platoon is split (Line 25) and the reference velocity is adjusted for the front subgroup (Line 26).
\item \textbf{Final Stability Verification (Consensus Check) (Lines 29-31):}\\  
After deciding on whether to proceed with the entire platoon or to split it, the algorithm performs a final stability (consensus) check via \texttt{VerifySplitStability} (Line 30). This step includes a 0.2-second consensus period to confirm that the front-rear grouping remains stable over time. Finally, the algorithm returns the optimized reference velocity along with the front (\(\mathcal{P}_f\)) and rear (\(\mathcal{P}_r\)) sub-platoons (Line 31).
\item \textbf{GetArrivalTimes Function (Lines 33-37):}\\  
The \texttt{GetArrivalTimes} function calculates the raw arrival times for the platoon. It computes:
\begin{itemize}
    \item The leader’s arrival time as \(t_{\text{lead}} = d_{\text{tl}} / v\) (Line 34).
    \item The last vehicle’s arrival time as \(t_{\text{last}} = (d_{\text{tl}} + L) / v\) (Line 36), where \(L\) is determined by \texttt{TotalPlatoonLength} (Line 35).
\end{itemize}

\item \textbf{CheckLeaderArrival Function (Lines 39-47):}\\  
The \texttt{CheckLeaderArrival} function refines the leader’s arrival time. It begins by computing the leader’s arrival time as \(t_{\text{arrival}} = d_{\text{tl}}/v\) (Line 40). If this computed time is less than \(t_g^{\text{start}}\), the function adjusts the velocity (by slowing down) so that the leader arrives closer to \(t_g^{\text{start}}\) (Line 42). Conversely, if the arrival time is greater than \(t_g^{\text{end}}\), it adjusts the velocity to ensure an earlier arrival (Line 44). The adjusted velocity is then clamped within the feasible bounds (\(v_{\min}\) and \(v_{\max}\)).


\item \textbf{SplitPlatoon Function (Lines 48-59):}\\  
The \texttt{SplitPlatoon} function manages the division of the platoon into front and rear subgroups. It starts by assigning the leader (vehicle \(p_1\)) in the front group (Line 49). Then, starting from the second vehicle (Line 50), it iteratively adds vehicles to the front group. For each added vehicle, the function checks, via \texttt{TrailingVehicleFeasible} (Line 52), whether the current reference velocity \(v_{\text{ref}}\) still allows the trailing vehicles to pass within the green window. If adding another vehicle would violate this condition, the loop breaks (Line 53). Finally, the rear subgroup is defined as those vehicles not included in the front group (Line 57)
\item \textbf{Computational Complexity:} \\
The algorithm operates in linear time \(O(n)\) relative to the number of vehicles, mainly due to the iterative loop in the splitting phase (Lines 50–56). All other operations run in constant time \(O(1)\). The overall space complexity is \(O(n)\), making the approach scalable for real-time traffic management systems.
\end{itemize}
\section{Evaluation}
\label{sec:evaluation}
The proposed \textit{green window advisory system} and \textit{platooning strategy} are evaluated based on key performance metrics such as \textbf{fuel consumption}, \textbf{traffic flow efficiency}, and \textbf{vehicle synchronization}. To illustrate the effectiveness of the proposed system, we compare two scenarios:  
\begin{enumerate}
    \item \textbf{With green window advisory and platooning}: Vehicles dynamically adjust their velocities to traverse signalized intersections efficiently.
    \item \textbf{Without these features}: Vehicles travel independently, leading to more frequent stops and higher energy consumption.
\end{enumerate}
All scenarios run until the leader completes two full laps and reaches its initial spawn point.

\begin{minipage}{\columnwidth}
 \hrule height 1pt
 \vskip 3pt
  \textbf{Algorithm 1: Pseudo Code for Platoon Coordination} 
  \vskip 3pt
  \footnotesize
  \begin{tabular}{r@{\quad}l}
    \midrule
  \label{algo:platoon_coordination}

  1.  & \textbf{Input:} \\
  2.  & \quad $d_{\text{tl}}$: Distance to the traffic light \\
  3.  & \quad $\mathcal{P}$: Ordered set of platoon vehicles (leader to tail) \\
  4.  & \quad $[t_g^{\text{start}}, t_g^{\text{end}}]$: Green window of the traffic light \\
  5.  & \quad $v_{\text{sat}}$: Saturation flow velocity \\
  6.  & \textbf{Output:} \\
  7.  & \quad $v_{\text{ref}}$: Reference velocity for platoon coordination \\
  8.  & \quad $\mathcal{P}_f,\,\mathcal{P}_r$: Front and rear subplatoons \\
  9.  & \textbf{Procedure:} \textsc{PlatoonCoordination}($d_{\text{tl}}, \mathcal{P}, [t_g^{\text{start}},t_g^{\text{end}}], v_{\text{sat}}$) \\
  10. & \quad \textbf{/* Initial Delay Check */} \\
  11. & \quad \textbf{if} \textsc{InitialDelayActive}() \textbf{then} \\
  12. & \quad\quad \Return $(v_{\text{current}}, \emptyset, \emptyset)$ \quad \Comment{Stabilization phase} \\
  13. & \quad \textbf{end if} \\
  14. & \quad $(v_{\min}, v_{\max}) \gets$ \textsc{GetFeasibleVelocities}($d_{\text{tl}}, t_g^{\text{start}}, t_g^{\text{end}}$) \\
  15. & \quad \textbf{/* Saturation Flow Verification */} \\
  16. & \quad \textbf{if} \textbf{not} \textsc{SaturationCheck}($\mathcal{P}, d_{\text{tl}},d_{\text{platoon}} ,t_g^{\text{start}}, t_g^{\text{end}}, v_{\max}$)  \\
  17. & \quad\quad $(\mathcal{P}_f, \mathcal{P}_r) \gets$ \textsc{SplitPlatoon}($\mathcal{P}$) \quad \Comment{Mandatory split} \\
  18. & \quad\quad $v_{\text{ref}} \gets$ \textsc{AdjustForSplit}($\mathcal{P}_f$) \\
  19. & \quad \textbf{else} \\
  20. & \quad\quad $v_{\text{cand}} \gets$ \textsc{OptimizeVelocity}($v_{\min}, v_{\max}$) \\
  21. & \quad\quad $(t_{\text{lead}}, t_{\text{last}}) \gets$ \textsc{GetArrivalTimes}($v_{\text{cand}}, d_{\text{tl}}, \mathcal{P}$) \\
  22. & \quad\quad \textbf{if} $t_{\text{lead}} \in [t_g^{\text{start}}, t_g^{\text{end}}]$ \textbf{and} $t_{\text{last}} \le t_g^{\text{end}}$ \textbf{then} \\
  23. & \quad\quad\quad \Return $(v_{\text{cand}}, \mathcal{P}, \emptyset)$ \quad \Comment{Entire platoon passes} \\
  24. & \quad\quad \textbf{else} \\
  25. & \quad\quad\quad $(\mathcal{P}_f, \mathcal{P}_r) \gets$ \textsc{SplitPlatoon}($\mathcal{P}$) \quad \Comment{Adaptive split} \\
  26. & \quad\quad\quad $v_{\text{ref}} \gets$ \textsc{AdjustForSplit}($\mathcal{P}_f$) \\
  27. & \quad\quad \textbf{end if} \\
  28. & \quad \textbf{end if} \\
  29. & \quad \textbf{/* Consensus period: 0.2 s */} \\
  30. & \quad \textsc{VerifySplitStability}() \\
  31. & \quad \Return $(v_{\text{ref}}, \mathcal{P}_f, \mathcal{P}_r)$ \\
  32. & \textbf{End Procedure} \\
  33. & \textbf{Function:} \textsc{GetArrivalTimes}($v, d_{\text{tl}}, \mathcal{P}$) \\
  34. & \quad $t_{\text{lead}} \gets d_{\text{tl}} / v$ \\
  35. & \quad $L \gets$ \textsc{TotalPlatoonLength}($\mathcal{P}$) \\
  36. & \quad $t_{\text{last}} \gets (d_{\text{tl}} + L) / v$ \\
  37. & \quad \Return $(t_{\text{lead}}, t_{\text{last}})$ \\
  38. & \textbf{End Function} \\
  39. & \textbf{Function:} \textsc{CheckLeaderArrival}($v, d_{\text{tl}}$) \\
  40. & \quad $t_{\text{arrival}} \gets d_{\text{tl}} / v$ \\
  41. & \quad \textbf{if} $t_{\text{arrival}} < t_g^{\text{start}}$ \textbf{then} \\
  42. & \quad\quad $v \gets \max(v_{\min}, d_{\text{tl}} / t_g^{\text{start}})$ \\
  43. & \quad \ElsIf{$t_{\text{arrival}} > t_g^{\text{end}}$} \\
  44. & \quad\quad $v \gets \min(v_{\max}, d_{\text{tl}} / t_g^{\text{end}})$ \\
  45. & \quad \textbf{end if} \\
  46. & \quad \Return \Call{Clamp}{$v, v_{\min}, v_{\max}$} \\
  47. & \textbf{End Function} \\
  48. & \textbf{Function:} \textsc{SplitPlatoon}($\mathcal{P}$) \\
  49. & \quad $\mathcal{P}_f \gets \{p_1\}$ \quad \Comment{Leader is always in front} \\
  50. & \quad \textbf{for} $i = 2$ \textbf{to} $|\mathcal{P}|$ \textbf{do} \\
  51. & \quad\quad $L_i \gets \operatorname{PlatoonLengthUpTo}(i,\mathcal{P})$ \\
  52. & \quad\quad \textbf{if}\ \textbf{not}\ $\operatorname{TrailingVehicleFeasible}(L_i, v_{\text{ref}})$\ \textbf{then} \\
  53. & \quad\quad\quad \textbf{break} \quad \Comment{Do not add more vehicles} \\
  54. & \quad\quad \textbf{end if} \\
  55. & \quad\quad $\mathcal{P}_f \gets \mathcal{P}_f \cup \{p_i\}$ \\
  56. & \quad \textbf{end for} \\
  57. & \quad $\mathcal{P}_r \gets \mathcal{P} \setminus \mathcal{P}_f$ \\
  58. & \quad \Return $(\mathcal{P}_f, \mathcal{P}_r) $\\
  59. & \textbf{End Function} \\
  \end{tabular}
   \vskip 5pt
  \hrule height 1pt
\end{minipage}

\vspace*{0.1cm}

\subsection{Trajectory Analysis and Platoon Splitting Behavior}

\begin{figure}[h]
    \centering
    \includegraphics[width=6cm, height=4cm]{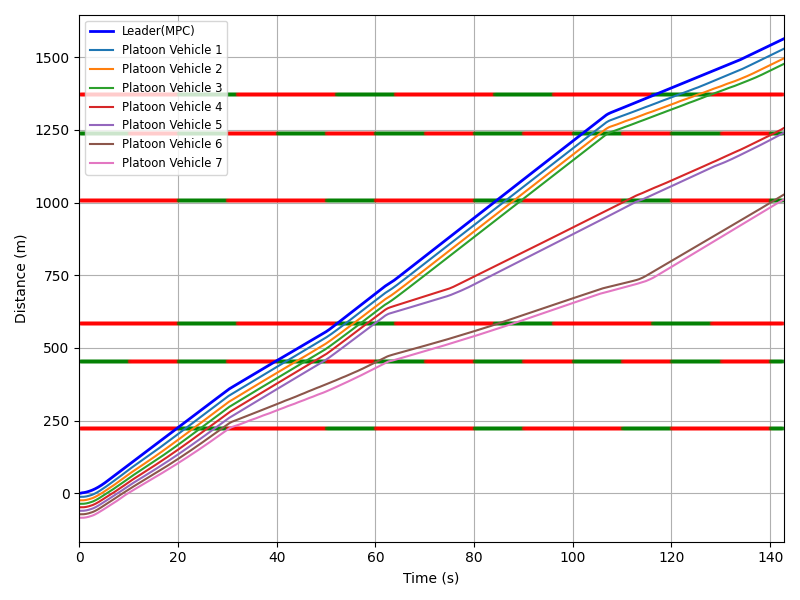}
    \caption{Vehicle trajectories illustrating platoon synchronization and dynamic platoon splitting decisions based on SPaT-driven reference velocities. Colored lines represent individual vehicle paths, while horizontal lines indicate intersection locations and corresponding signal phases.}
    \label{fig:3er_Split}
\end{figure}

Figure~\ref{fig:3er_Split} illustrates the synchronized movement of the platoon throughout the circular route. The colored curves represent vehicles numbered from 0 to 7, while the horizontal lines indicate signalized intersections. The color of these lines corresponds to the current signal phase, providing a visual representation of traffic signal states along the route. The simulation starts in Corridor 3 and follows the sequence:
\begin{itemize}
    \item Corridor 3: 0 – 220 m
    \item Corridor 1: 220 – 420 m
    \item Corridor 2: 420 m – 580 m
    \item Corridor 3 (Re-entered): 580 m onward till 800 m
\end{itemize}

\begin{figure}[h]
    \centering
    \includegraphics[width=7cm, height=5cm]{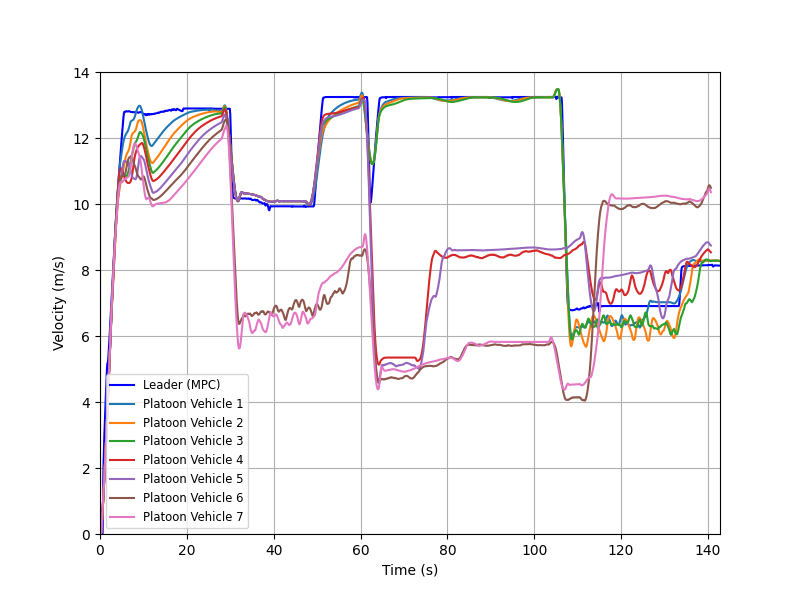}
    \caption{Detailed analysis of platoon splitting behavior at intersections under SPaT-driven velocity control. Key events include the initial accordion effect (5–10 s), first platoon split (30 s), and second platoon split (60–70 s). The trajectories highlight adaptive platoon formation, robust vehicle coordination, and effective handling of temporary communication losses (DSRC range limitations).}
    \label{fig:platoon_velocity}
\end{figure}

Since the route is circular, vehicles traverse the same corridors multiple times, dynamically adjusting their speed based on upcoming signal states. Key observations from this scenario include:
\begin{itemize}
    \item \textbf{Parallel trajectories}, indicating synchronized movement across all vehicles.
    \item \textbf{Minimal stops at intersections}, reducing travel time and idling.
    \item The \textbf{MPC-controlled platoon leader} effectively regulates velocity to ensure smooth transitions and optimize split decisions.
    \item Figure~\ref{fig:3er_Split} demonstrates the formation of \textbf{multiple platoons}, each with its leader, to navigate intersections based on reference velocity. Figure~\ref{fig:platoon_velocity} provides further insights into the dynamic platoon behavior.
\end{itemize}
\textbf{Accordion Effect(5-10 s):}  
\begin{itemize}
    \item As explained in Section~\ref{sec:algo}, during the first five seconds, no computation of the reference velocity is performed, and no V2X communication takes place. This initial period is dedicated to forming the platoon, allowing vehicles to establish communication links and form a cohesive group. Once communication is established, vehicles begin exchanging information to maintain the desired inter-vehicle gaps. This coordination phase often triggers what we refer to as the "accordion effect", a phenomenon where vehicles dynamically decelerate and adjust their spacing (much like the compression and expansion of an accordion) to maintain safe and consistent gaps as the platoon stabilizes.
\end{itemize}

\textbf{First Split (30 s):}  
\begin{itemize}
    \item Platoon Vehicle 6-7 \textbf{deviate from the main group}, taking a different path as a \textbf{sub-platoon} from corridor 2 to corridor 3 one cycle after the original platoon.
    \item This decision is based on the leader's \textbf{reference velocity} calculated using SPaT data, ensuring efficient traversal of upcoming green signals.
\end{itemize}

\textbf{Second Split (60–70 s):}  
\begin{itemize}
    \item Leader decides to split, and vehicles 4 and 5 form a \textbf{sub-platoon}. SPaT is temporarily unavailable for the sub-platoon leader due to DSRC range limitations (as explained in Section~\ref{sec:algo}).
    \item During this period, they attempt to \textbf{maintain a constant velocity} to avoid sudden acceleration or braking.
    \item Once they re-enter the DSRC range, they receive updated SPaT data and adjust their velocities accordingly.
\end{itemize}

These results highlight the system's ability to \textbf{adaptively manage platoon formations}, even when vehicles temporarily lose communication with traffic signals and underline several key advantages of the proposed system:
\begin{itemize}
    \item \textbf{Efficient traffic flow:} Vehicles avoid unnecessary stops and adjust speed dynamically to traffic light timings.
    \item \textbf{Reduced braking and acceleration:} Synchronization minimizes fuel-consuming speed fluctuations.
    \item \textbf{Efficient energy utilization:} The system optimizes velocity trajectories to improve fuel efficiency.
    \item \textbf{Adaptive platoon coordination:} Vehicles can split and reform into sub-platoons as necessary, ensuring robustness under DSRC communication constraints.
\end{itemize}

\begin{figure}[h]
    \centering
    \includegraphics[width=6cm, height=4cm]{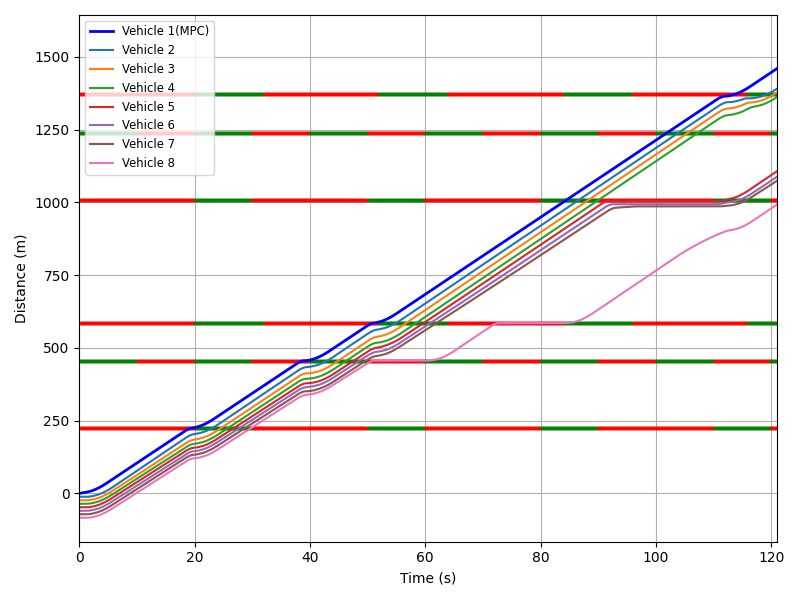}
    \caption{ Vehicle trajectories without coordinated platooning and SPaT-based green window advisory. The irregular and abrupt changes in slope illustrate frequent stops and accelerations at intersections, highlighting inefficiencies caused by independent vehicle decisions and lack of predictive velocity control.}
    \label{fig:without_green}
\end{figure}

In contrast, the trajectories in the scenario without green advisory and platooning reveal irregular movement patterns, as seen in Figure~\ref{fig:without_green}. Vehicles operate independently, reaching the maximum allowable speed of 50~km/h but frequently encountering red lights. These frequent stops and starts result from the absence of coordinated platooning and green window advisory. Key observations from this scenario include:
\begin{itemize}
    \item Abrupt changes in slope or pauses in the trajectories, indicating frequent stops at red lights.
    \item Independent decision-making by vehicles results in less efficient roadway utilization and higher travel times.
    \item Increased energy consumption due to frequent stops and accelerations.
\end{itemize}

\subsection{Fuel Consumption Analysis}
\begin{figure}[h]
    \centering
    \includegraphics[width=6cm, height=4cm]{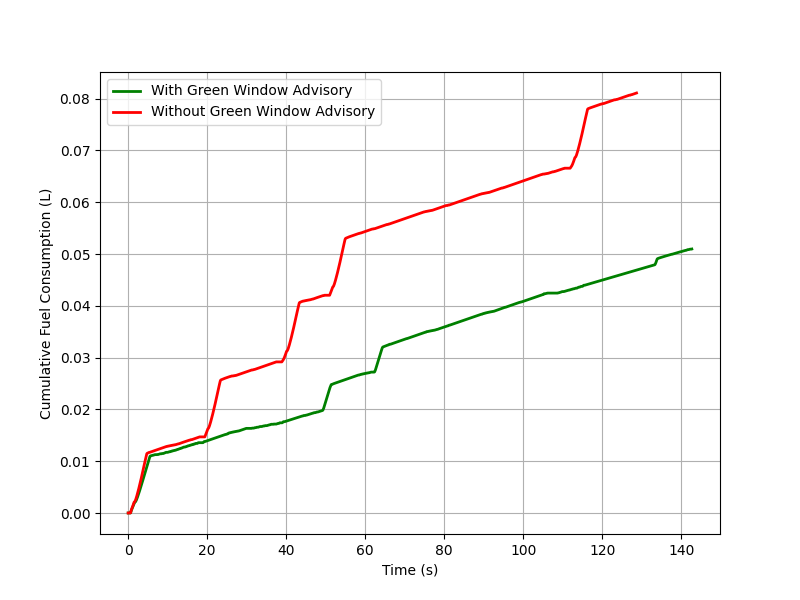}
    \caption{Cumulative fuel consumption comparison for the MPC-controlled platoon leader vehicle under scenarios with and without SPaT-based green window advisory. The SPaT-based advisory system demonstrates smoother velocity profiles, resulting in lower and steadier fuel usage.}
    \label{fig:FuelCon_compare}
\end{figure}

The fuel consumption analysis focuses on the platoon leader controlled by the MPC. The comparison highlights significant differences between the two scenarios. As shown in Figure~\ref{fig:FuelCon_compare}, cumulative fuel consumption increases steadily and gradually with green advisory, reflecting smoother driving conditions and optimized velocity profiles. The green advisory system enables the platoon leader to make smoother velocity adjustments, thereby reducing unnecessary fuel usage. Without a green advisory system, cumulative fuel consumption rises sharply due to frequent stops and accelerations caused by red lights. This behavior results in higher energy consumption. 
\begin{table}[h]
    \centering
    \large
    \resizebox{\columnwidth}{!}{
    \begin{tabular}{|l|c|c|}
        \hline
        \textbf{Scenario} & \textbf{Leader Fuel Consumption (L)} & \textbf{Total Platoon Fuel Consumption (L)} \\
        \hline
        \textbf{No Green Window Advisory}  & 0.08 & 0.63 \\
        \textbf{With Green Window Advisory} & 0.05 & 0.37 \\
        \hline
        \textbf{Fuel Savings (\%)}  & \textbf{37.5\%} & \textbf{41.2\%} \\
        \hline
    \end{tabular}
    }
    \caption{Fuel consumption comparison for the platoon leader and the entire platoon under scenarios with and without SPaT-based green window advisory, highlighting substantial fuel savings achieved through coordinated platooning}
    \label{tab:fuel_comparison}
\end{table}
Table~\ref{tab:fuel_comparison} demonstrates the effectiveness of the SPaT-driven green window advisory system in reducing fuel consumption. The leader achieves \textbf{37.5\% fuel savings}, benefiting from optimized velocity control that minimizes unnecessary braking and acceleration at intersections.
For the entire platoon, \textbf{fuel savings reach 41.2\%}, reflecting smoother traffic flow and reduced speed fluctuations. Unlike the no-green-advisory scenario, where the platoon leader consumes 0.08 L, and the total platoon consumption is 0.63 L, implementing the green window advisory reduces these values to 0.05 L and 0.37 L, respectively. The significant savings demonstrate the impact of the coordinated platoon movement and SPaT-based speed optimization, which prevent stop-and-go driving at intersections. 
\section{Future Works}
\label{sec:future_works}
The proposed advisory algorithm offers a structured method for urban platooning, reducing stops and improving fuel efficiency through SPaT-based synchronization. It's a 0.2-second decision buffer that improves stability but assumes perfect signal timing, an assumption that may not hold with adaptive or malfunctioning lights. Incorporating probabilistic signal predictions could enhance robustness. Extending the system to multi-corridor coordination using Cellular V2X (C-V2X) could better address complex urban traffic. Communication reliability also remains a challenge; packet loss or delays may disrupt platoon coordination. Adaptive algorithms are needed to manage real-world uncertainties such as erratic human drivers, unexpected stops, and dynamic traffic conditions. Future work will also explore jerk-limited trajectories to improve passenger comfort and adopt more accurate fuel consumption models. Additionally, addressing V2X cybersecurity, through robust communication protocols and intrusion detection, will be critical for safe deployment.

\section{Conclusion}
\label{sec:conclusion}
This paper presents a centralized platooning framework that combines V2X communication, SPaT-based green window advisory, and nonlinear MPC for fuel-efficient urban driving. By dynamically adjusting reference velocities and supporting adaptive platoon splitting, the system reduces idling, improves intersection throughput, and cuts fuel consumption by up to 40\%. Validated in CARLA, the framework operates at 10 Hz in quasi-real time and will be open-sourced to enable real-world deployment and foster further research in sustainable urban mobility.

\section{Acknowledgment}
\label{sec:acknowledgment}
This paper was created in the Country 2 City - Bridge project of the German Center for Future Mobility, which is funded by the German Federal Ministry for Digital and Transport.

{\small
\bibliographystyle{IEEEtran}
\bibliography{references}
}

\end{document}